\documentclass[10pt,twocolumn,letterpaper]{article}

\usepackage{cvpr}      

\usepackage[dvipsnames]{xcolor}

\definecolor{cvprblue}{rgb}{0.21,0.49,0.74}
\usepackage[pagebackref,breaklinks,colorlinks,citecolor=cvprblue]{hyperref}

\usepackage{url}
\usepackage{graphicx}
\usepackage{multirow, makecell}
\usepackage{booktabs}
\usepackage{subcaption}
\usepackage{comment}
\usepackage{xspace}

\usepackage{amsmath,amsfonts,bm}

\def\eqref#1{equation~\ref{#1}}

\def\1{\bm{1}}

\def\va{{\bm{a}}}

\def\vh{{\bm{h}}}
\def\vi{{\bm{i}}}
\def\vj{{\bm{j}}}

\def\vo{{\bm{o}}}

\def\vt{{\bm{t}}}
\def\vu{{\bm{u}}}
\def\vv{{\bm{v}}}
\def\vw{{\bm{w}}}
\def\vx{{\bm{x}}}

\def\vz{{\bm{z}}}

\DeclareMathAlphabet{\mathsfit}{\encodingdefault}{\sfdefault}{m}{sl}
\SetMathAlphabet{\mathsfit}{bold}{\encodingdefault}{\sfdefault}{bx}{n}

\def\va{{\bm{a}}}

\def\vh{{\bm{h}}}
\def\vi{{\bm{i}}}
\def\vj{{\bm{j}}}

\def\vo{{\bm{o}}}

\def\vt{{\bm{t}}}
\def\vu{{\bm{u}}}
\def\vv{{\bm{v}}}
\def\vw{{\bm{w}}}
\def\vx{{\bm{x}}}

\def\vz{{\bm{z}}}

\newcommand{\gray}[1]{\textcolor{gray}{{#1}}}

\newcommand{\ours}{Mirasol3B \xspace}

\def\paperID{7514} 
\def\confName{CVPR}
\def\confYear{2024}

\title{Mirasol3B: A Multimodal Autoregressive Model for Time-Aligned and Contextual Modalities}

\author{AJ Piergiovanni\\
Google DeepMind\\
\and
Isaac Noble\\
Google Research \\
\and
Dahun Kim  \\
Google DeepMind\\
\and
Michael S. Ryoo  \\
Google DeepMind\\
\and
Victor Gomes\\
Google Research
\and
Anelia Angelova \\
Google DeepMind\\
}

\begin{document}
\maketitle
\begin{abstract}

One of the main challenges of multimodal learning is combining multiple heterogeneous modalities, e.g., video, audio, and text.
Video and audio are obtained at much higher rates than text and are roughly aligned in time. They are often not synchronized with text, which comes as a global context, e.g. a title, or a description. Furthermore, video and audio inputs are of much larger volumes, and grow as the video length increases, which naturally requires more compute dedicated to these modalities, and makes modeling of long-range dependencies harder. 
We here decouple the multimodal modeling, dividing it into separate autoregressive models, processing the inputs according to the characteristics of the modalities.
We propose a multimodal model, consisting of an autoregressive component for the time-synchronized modalities (audio and video), and an autoregressive component for the context modalities which are not necessarily aligned in time but are still sequential. To address the long-sequences of the video-audio inputs, we further partition the video and audio sequences in consecutive snippets and autoregressively process their representations. To that end, we propose a Combiner mechanism, which models the audio-video information jointly,  
producing compact but expressive representations. This allows us to scale to 512 input video frames without increase in model parameters. 
Our approach achieves the state-of-the-art on multiple well established multimodal benchmarks. 
It effectively addresses the high computational demand of media inputs by learning compact representations, controlling the sequence length of the audio-video feature representations, and modeling their dependencies in time.






\end{abstract}


\section{Introduction}

\begin{figure}[t]
\begin{center}
\includegraphics[width=0.97\linewidth]{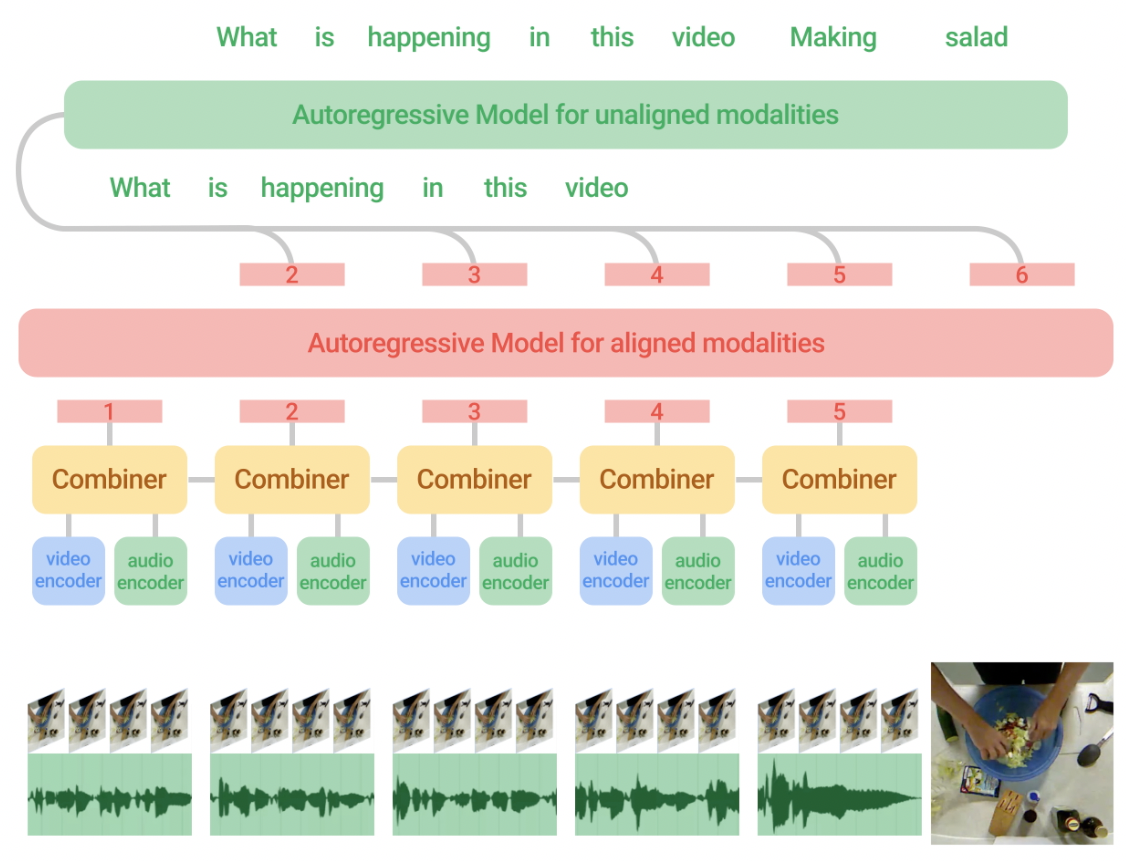}
\end{center}
\caption{
Autoregressive learning of time-aligned video and audio modalities, in time, 
and decoupling from the autoregressive text modeling  
allows for more effective multimodal models at smaller sizes and leads to scaling to longer videos.
 }
\label{fig:teaser}
\end{figure}
Multimodal models aim to combine the signals from multiple varied sources, which makes them both universal and useful for practical applications. 
However, these modalities have diverse characteristics and are challenging to combine uniformly by a single model.
 For example, video and text have disparate sampling rates: a video has many frames per second, but text or other similar types of global context, e.g., a description or title, 
 can be provided once per video, or asynchronously to the video. 
Video also 
 takes a larger portion of the input.
 At the same time, video and audio are naturally co-occurring and appear (almost) synchronously.
 They are roughly aligned and complementary. This co-occurrence in time 
 can contribute to their joint learning and serve as a rich self-supervisory learning signal, applied more frequently than global text signals.
 So, ideally, these modalities need to be processed by differently-synchronized model components, which process more adequately inputs of different frequencies and allocate more parameters to the more abundant modalities. 

 
 

 Following the success of large language models, where text input sequences are processed autoregressively, many recent multimodal models reuse the autoregressive text models, feeding in other modalities, e.g., as embeddings, 
 ~\citep{pali,wang2022git,piergiovanni2022answer,li2021align,li2022blip,flamingo}, or by tokenizing the visual inputs to be processed together with the text token sequence~\citep{wang2022image,CM3,CM3Leon,gato,merlot-reserve}). 
 %
 %
%
%
However, the imbalance of the information volume is large and models which are well suited to encode/decode text sequences process only highly compressed image or video features~\citep{flamingo,attention-bottlenecks}.  For example, the Flamingo model~\citep{flamingo}, subsamples the video features significantly, dedicating only about 1\% of the parameters to the image and video inputs, leaving the rest for text processing. Alternatively, methods that process the video, running each frame independently through an encoder or a tokenizer, can process only a limited number of frames
~\citep{videococa, dynpretr}. 
For longer inputs, these representations are insufficient to properly represent the modalities, 
which inherently limits 
the ability to model fine-grained or long-range dependencies.

%


 We here propose an audio-video-text multimodal model, where we decouple the autoregressive modeling into a component for time-aligned  modalities, e.g., audio and video, which are processed in time, autoregressively, and an autoregressive component for non-time-aligned contextual modalities e.g., text (\cref{fig:teaser}). Cross-attention weights coordinate the learning between these components. 
 This decoupling allows for better parameter distribution within the model, allocating sufficient capacity for the media modalities (video and audio), and leads to smaller models overall.
 Furthermore, we partition the time-aligned modalities into time segments, where audio-video representations are jointly learned before modeling their features autoregressively in time. 
 To that end, we introduce a joint feature learning mechanism for audio and video, called  the Combiner, which fuses their features and produces a more compact representation.
 We extract low level spatio-temporal representation from the raw media inputs in order to capture the dynamic nature of videos 
 and combine it with audio features within concurrent timesteps.  
  The Combiner effectively balances the need for efficient audio+video representations and ones which are expressive enough to preserve the media content.
  It sufficiently represents the events and activities in the videos and other concurrent modalities and can be handled by subsequent autoregressive models, which allows for learning of long-range dependencies. 
  Our model enables consuming multimodal inputs at different rates and scales well with longer videos.
 %
  %
   %
%
Our contributions are: 
\begin{itemize}
    \item An autoregressive multimodal model, subdividing learning into autoregressive modeling for time-aligned media modalities and non-time-aligned contextual modalities.  
    \item  Joint feature representation learning via the Combiner to balance the learning of efficient video+audio representations which are also sufficiently expressive. 
    \item We demonstrate learning with 128-512 frames  without increase in model parameters. This is in contrast to prior multimodal models that use 8 or 32 frames~\citep{videococa,gao2023mist}.
\end{itemize}

Our model outperforms the state-of-the-art on multiple benchmarks, with large margins on audio-video-text datasets and on long video datasets.
%
%


\section{Related work}

Architectures for video-language understanding commonly use a joint transformer, where video inputs are fed in together with text tokens and processed autoregressively~\citep{fu2021violet,merlot}). This is often accomplished with tokenizing the visual inputs. 
%
Video-text pretraining approaches ~\citep{HowTo100M,miechEnd-to-end,internvideo,omniVL,li2023umt,alpro} use  masked token modeling and reconstruction~\citep{fu2021violet}, 
masking with cross-attention on multimodal inputs~\citep{flamingo}, 
or
contrastive learning
~\citep{omniVL,mPLUG-2,merlot,weakly_sup}. 
Visual synthesis models 
have extensively used autoregressive models, by learning to generate pixel-wise predictions 
~\citep{PixelCNN}, or by learned discrete tokens from images or videos~\citep{nuwa,VideoGPT,godiva}. 
In other models, encoder-decoder or decoder-only architectures extend an image-text model to a video-text one 
~\citep{wang2022git,MaMMUT,dynpretr,videococa}, where video is processed by individual frames which are then combined.
Some architectures instead extract full video signals (typically as embeddings) before feeding them to the model~\citep{mPLUG-2}. Another option is to attach a projection or re-tokenization layers e.g., as in Perceiver in Flamingo~\citep{flamingo}, to reduce the amount of visual tokens added to the model.
Our approach  
differs substantially, as the media input features have a specifically designed component to learn them jointly and in time, 
producing more abstract representations, suitable for modeling long videos.

\begin{figure*} 
\begin{center}
\includegraphics[width=0.9\linewidth]{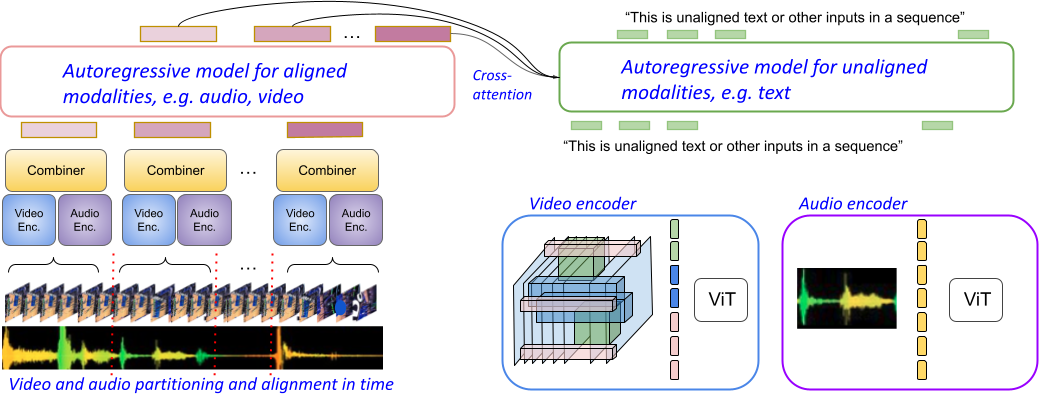}
\end{center}
\caption{
The \ours model architecture consists of an autoregressive model for the time-aligned modalities, such as audio and video, which are partitioned in chunks 
(left) and an autoregressive model for the unaligned context modalities, which are still sequential, e.g., text 
(right). This allows adequate computational capacity to the video/audio time-synchronized inputs, including processing them in time autoregressively, before fusing with the autoregressive decoder for unaligned text (right). Joint feature learning is conducted by the Combiner, balancing the need for compact representations and allowing sufficiently informative features to be processed in time.}
\label{fig:arch}
\vspace{-1em}
\end{figure*}





Multimodal audio-video-text models have also gained popularity~\citep{UAVM,merlot-reserve,CAV-MAE,Huang2022mavil,sun2023FineGrained}: 
UAVM~\citep{UAVM} propose joint learning of audio and video by building invariant transformer module which can be reused by either signal. 
Multimodal Transformer~\citep{mmtrasf} proposes cross-attention mechanisms, for cross-modal learning on all pairs of video-audio-text data, which~\citet{STREAMULT} extends to longer sequences.
\citet{merlot-reserve} demonstrate joint multimodal audio-video-text learning but only aligning text and audio. 
\citet{CAV-MAE} use contrastive audio-video learning, whereas 
\citet{Huang2022mavil} use masked autoencoder for audio-video learning. 
Both approaches tokenize the audio video inputs independently in 2D patches which are used for further processing. 
Contrastive learning for audio-video signals, leveraging the time-alignment between them~\citep{AVLnet,korbar2018cooperative} and 
 audio-video late fusion are also common~\citep{avfusion}.

Our work is related to long-form video understanding~\citep{towards-long-form,long-form}. 
Long-form videos have been handled by hierarchical feature learning e.g., the Temporal Window Attention~\citep{long-form} where dependencies are learned locally and then further propagated to higher level cross-attention modules. 
\citet{kumar2023hiervl} propose contrastive learning at different hierarchical levels. \citet{gao2023mist} segment videos then pool their features into a small representation.
Memory-augmented model for long videos 
~\citep{memvit} accumulate prior context in learnable `memory', to be referenced at each step of learning. 
Our work contributes by proposing a balanced approach of locally learning important features, jointly within the modalities.

\section{Approach}
\label{sec:main}


Autoregressive models are powerful generative models that are well suited for data which appears in a sequence, modeling the probability of the current value, conditioned of previous ones. 
Video and audio information is sequential but also roughly time-synchronized. At the same time, other modalities e.g., text, might be provided globally per video as context and applied to the full video rather than to specific parts\footnote{Text, e.g., ASR, might appear concurrently with audio/video and can contribute to improved understanding of the video content. We leave this to future work.}. 
To address the challenges of modeling diverse multimodal inputs, we propose to subdivide the autoregressive modeling by learning separate autoregressive models: one for the time-aligned modalities (audio-video), \cref{sec:autoregressive}, and another one for modalities which are not necessarily aligned in time but are still sequential, \cref{sec:autoregressive_text}. Learning across these is coordinated by 
cross-attention mechanisms, where here the audio+video inputs are allocated a lot more parameters and are properly modeled in time. 
 A learning module, called the Combiner (\cref{sec:Combiner}), combines the lower-level signals from video/audio snippets. Here information is processed spatio-temporally, extracting features particularly relevant to dynamic changes in the inputs. 
%



\begin{figure}[t]
\begin{center}
\includegraphics[width=0.99\linewidth]{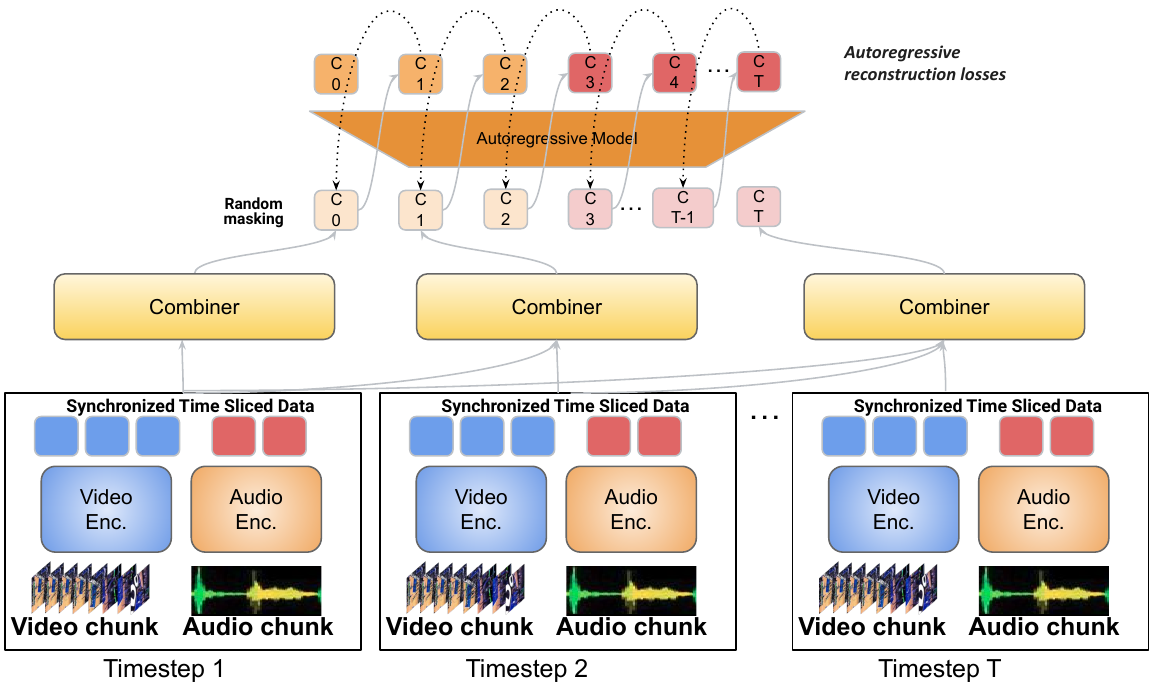}
\end{center}
\vspace{-3mm}
\caption{Autoregressive modeling of video and audio in time.}
\label{fig:ar_arch}
\end{figure}

\begin{figure}[t]
\begin{center}
\includegraphics[width=0.95\linewidth]{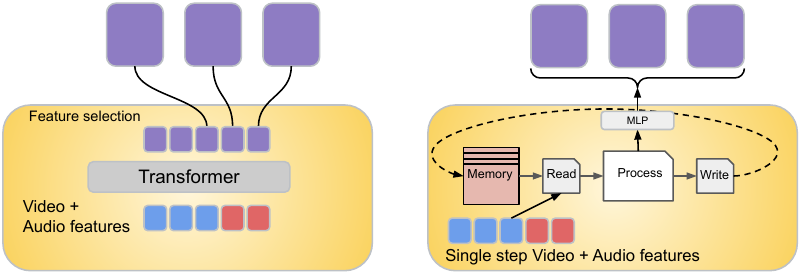}
\end{center}
\vspace{-4mm}
\caption{Combiners: Transformer Combiner (left): all features are input to the transformer, a smaller number of m features are selected as combined features.
  TTM Combiner (right): uses the TTM mechanism to store a memory and compute the m combined features for each time step.
This process is repeated for each step.}
\label{fig:ar_arch_comb}
\vspace{-5mm}
\end{figure}

\textbf{Architecture overview.} 
At a high level, the architecture consists of two main learning components (\cref{fig:arch}): 
%
%
The first one is an autoregressive component which is designed to process (almost) synchronized multimedia inputs e.g., video+audio and combine their inputs in time (\cref{fig:ar_arch}). 
In order to process the video and audio signals, and to accommodate longer video/audio inputs, they are partitioned into smaller chunks (roughly synchronized in time) for which a joint audio-visual representation is learned via the Combiner as described below (\cref{fig:ar_arch_comb}). 
The second component processes the context, or the signals not aligned in time, such as global text information, 
which are often still sequential. It is autoregressive as well, and  uses the combined latent space as cross-attention inputs. 

\textbf{Model inputs.} We have an input video sequence of $N$ frames $\vv=\{ \vv^f_1, \vv^f_2, \ldots \vv^f_N \}$, and audio wave signal of $M$ timesteps $\va=\{ \va^f_1, \va^f_2, \ldots \va^f_M \}$, where the audio signal is captured during the duration of the video and corresponds to the given video input. Additionally we have an input text sequence $\vt=\{ \vt^f_1, \vt^f_2, \ldots \vt^f_P \}$, which is related to the video/audio and might vary according to the task e.g. it can be a description, a question-answer pair, meta information.

\textbf{Partitioning of the media inputs.} In order to process the video sequence efficiently and to learn the correlation of features in time, we partition the input video into into $T$ non-overlapping segments or chunks, with $\vv_t$ and $\va_t$ denoting the video and audio input per segment (let $K=N/T$). Here each chunk captures all input data between two timestamps (i.e., video and audio snippets), as follows: 
\begin{equation}
\label{eq:1}
    \underbrace{\vv^f_1,\vv^f_2,\ldots, \vv^f_{K},}_{\vv_1}
    \underbrace{\vv^f_{K+1},\vv^f_{K+2},\ldots, \vv^f_{2K},}_{\vv_2} \dots 
    \underbrace{\vv^f_{(T-1)K+1},\ldots, \vv^f_N,}_{\vv_T}
\end{equation}
Thus the video is represented by its chunks instead, $\vv=\{ \vv_1, \vv_2, \ldots \vv_T \}$, and more specifically latent features will be extracted from each chunk to represent the video (as described in~\cref{sec:tube-vit}).
A similar partitioning is done for the audio signals, where they are partitioned in $T$ chunks to be synchronized in time to the video chunks,  $\va=\{ \va_1, \va_2, \ldots \va_T \}$.  Here too we assume that audio features will be extracted from the raw audio signal,~\cref{sec:tube-vit}. 

\subsection{Audio/video features}
\label{sec:tube-vit}

\textbf{Video features.} Prior models captured video information at individual sparsely sampled frames, which lacks the temporal information essential to video understanding and which might miss dynamic events. Alternatively, 3D convolutions~\citep{nuwa}, sparse 3D tubes~\citep{piergiovanni2022tubevit} and others learn spatio-temporally, which can capture key dynamic changes in the videos.
We expand on these ideas and extract sparse 3D tubes~\citep{piergiovanni2022tubevit} from the videos which span all 3 dimensions of the video snippet starting at various locations. The sparse 3D tubes, together with standard 2D patches are processed via a ViT encoder. Rather than applying offsets, 
the tubes are applied at the beginning of the snippet.  
Per video chunk $t$ we denote $\bm{\hat{v}}_t$ as the time-aligned features for this chunk, and thus  
$\bm{\hat{v}}=\{\bm{\hat{v}}_1, \bm{\hat{v}}_2, \ldots, \bm{\hat{v}}_T)$ are the  time-aligned video representations for the whole video. 

\textbf{Audio features.} Audio inputs arrive at a predefined frequency and can be processed in various ways. We here 
represent the audio as a spectrogram. The spectrogram is created so that the time bands match the 25 
frames per second used in the videos, and thus can easily be split into snippets aligned with the video. The spectrogram for each snippet is processed by a ViT model, after an audio input projection layer. The ViT backbone is the same as the one used for video features. Reusing the visual component is previously shown to be advantageous~\citep{UAVM}. 
Similarly to above, we denote $\bm{\hat{a}}_t$ to be the audio feature per chunk $t$ and $\bm{\hat{a}}=\{\bm{\hat{a}}_1, \bm{\hat{a}}_2, \ldots, \bm{\hat{a}}_T)$ for the full video.

\subsection{Modality Combiner}
\label{sec:Combiner}

The task of the Combiner module is two-fold: 1) to combine the video (and audio) features at a specific snippet of time, learning their joint representation and 2) 
effectively compress the representation from each video/audio snippet, which allows our model to scale to longer videos. 



When partitioning the inputs, the features for each modality, video and audio in this case, are (roughly) time-aligned latent features $\bm{\hat{v}}=\{\bm{\hat{v}}_1, \bm{\hat{v}}_2, \ldots, \bm{\hat{v}}_T)$ and $\bm{\hat{a}}=\{\bm{\hat{a}}_1, \bm{\hat{a}}_2, \ldots, \bm{\hat{a}}_T)$, 
where the maximum timestamp for any data incorporated into $\bm{\hat{v}}_t$ or $\bm{\hat{a}}_t$ is less than the minimum timestamp of any data incorporated into $\hat{\vv}_{t+1}$ or $\hat{\va}_{t+1}$. Explicitly $\bm{\hat{v}}_t$ is composed of $f$ features of size $d$ giving it a shape of $(f,d)$ and $\bm{\hat{a}}_t$ is composed of $s$ features also of size $d$ with shape $(s,d)$. The role of the Combiner is to map such time-aligned modal latent features into a smaller set of shared latent features. Specifically let $\bm{u} = \{\bm{u}_1, \bm{u}_2, \ldots, \bm{u}_T\}$ where $\bm{u}_t=(\bm{\hat{v}}_t,\bm{\hat{a}}_t)$ having size $(n,d)$ and $n=f+s$ be the set of all time-aligned features from all modalities. The Combiner then maps $\bm{u}$ to a lower dimensional latent feature space $\bm{x} = \{\bm{x}_1, \bm{x}_2, \ldots, \bm{x}_T\}$ where $\bm{x}_t$ has shape $(m,d)$ where $n >> m$.

Since the Combiner is reducing the dimensionality of video+audio features, it can effectively use all features in the sequence, not only the ones per chunk. 
However, since the features produced by the Combiner are going to be used in the sequential autoregressive modeling of video/audio, it is important for the Combiner to not break causality thus:
\begin{equation}
\bm{x}_t=\text{Combiner}(\bm{u}_1,\bm{u}_2, \ldots, \bm{u}_{t})
\label{eq:Combiner}
\end{equation}
 We utilize two different architectures for the Combiner,  a standard Transformer one and a memory based one, based on the Token Turing Machines~\citep{ttm}, to reduce memory. 

\paragraph{Causal Transformer Combiner.}

We explore a straightforward version of the Combiner, which consists of a standard Transformer model, here of $R$ layers (here, $R=8$). For each step $t$ it maps the original set of features $\vu_t$ to $\vx_t$ where $\vx_t$ is of much lower dimensionality, i.e., effectively reducing the number of tokens (here $m=32$) (\cref{fig:ar_arch_comb}). 
The inputs to the Combiner are the latent features of the video and audio, which are concatenated before being fed to the Combiner.
We here specifically implement a causal version of the transformer as it masks out inputs from future timestamps (i.e., $> t$). 
 The attention mechanism of the transformer is modified to mask features at the time-chunk level as described later in~\cref{sec:timeattn} (using~\cref{eq:timeattn}), thus all features from $\vu_t$ and the previous time steps are used to compute each output feature in $\vx_t$ as in~\cref{eq:Combiner}. This effectively applies attention mechanisms to all the modality inputs jointly while respecting causality.

\paragraph{Token Turing Machine Combiner.}

Token Turing Machine (TTM) \citep{ttm} is a recurrent sequential model with Transformers and token-based operations. It maintains an external `memory' $M_t$ as a set of features, and updates it at every time step by reading and writing.
Given inputs $\vu_t$ at each time step, it first `reads' features to be processed, from input features as well as memory features. Such features, $\vz_t$, are passed to the `processor', which is implemented as a standard Transformer, generating a set of intermediate output features $\vo_t$. These intermediate outputs are then used to update $M_t$ (i.e., memory `write') as well as to produce the final output $\vx_t$.
\begin{align}
    \vz_t &= \text{Read}(\vu_t, M_t) \\
    \vo_t &= \text{Process}(\vz_t) \\
    M_{t + 1} &= \text{Write}(M_t, \vo_t, \vu_t) \\
    \vx_t &= \text{Output}(\vo_t)
\end{align}

The key idea is to make the TTM processor generate the outputs by utilizing memory $M_t$ instead of the entire history of features $\{\vu_1, \dots, \vu_{t-1}\}$. Once trained, the differentiable read and write operations are optimized to maintain $M_t$ so that it stores important features from the previous time steps $\{\vu_1, \dots, \vu_{t-1}\}$ and updates it, at every step.

We implement TTM as the Combiner module to sequentially combine $\vu$. The function `Process' is implemented with a standard Transformer with layers of multi-head self-attention and MLPs. The functions `Read', `Write', and `Output' are implemented with TokenLearner \citep{ryoo2021tokenlearner_neurips} (which is similar to Perceiver \citep{perceiver} and attention pooling \citep{lee2019set}). 
Note that we are able to separately control the number of features in the memory as well as the number of `Output' function features, allowing efficient Combiner computation and feature generation.

The key advantage of the TTM Combiner is its utilization of memory features to sequentially process $\vu_t$. The number of such memory features are much smaller than the total number of history features ($\{\vu_1, \dots, \vu_{t-1}\}$) in general (e.g., 256 vs. $\sim$10k).
This not only makes TTM a natural fit for the model, but also reduces the total time complexity of the TTM Combiner to be constant with respect to $t$, instead of $O(t)$ or $O(t^2)$ in Transformers. We observe that the TTM Combiner saves memory in both training and inference, using about 30\% less memory and reduces the runtime by about $18\%$.


\subsection{Time-Aligned Video/Audio Autoregressive Modeling}
\label{sec:autoregressive}


We describe the autoregressive modeling of time-aligned video and audio. 
%
%
%
We apply autoregressive modeling strategy where we condition video/audio representations corresponding to a time interval on feature representations from previous time intervals. These representations are learned jointly by the Combiner, as described~\cref{sec:Combiner}.
As mentioned, the video is first partitioned in $T$ smaller video snippets. Each of the snippets itself can be of size 4-64 frames (overlap is possible but currently not used). We extract spatio-temporal information into latent video features $\bm{\hat{v}}_t$ and audio features $\bm{\hat{a}}_t$ from the same video partition, apply Combiner to produce $x_t$. 
The feature representations per video chunk $x_t$ are then fed sequentially to the autoregressive model, where at each step we reconstruct the features from the previous step, conditioned on the prior inputs and the latent vector $\vh$ which corresponds to the latent representations learned within the autoregressive model: 
\begin{equation}
p(\vv,\va) = \prod_{t=1}^T p(\vv_{t+1},\va_{t+1}|\vh_t)p(\vh_t|\vx_t)p(\vx_t|\vv_t,\va_t)
\label{eq:mm}
\end{equation}
where  $\{ \vv_1, \vv_2, \ldots \vv_T \}$, and $\{ \va_1, \va_2, \ldots \va_T \}$ are the feature representations from the video and audio, $p(\vx_{t}|\vv_{t},\va_{t})$ is estimated by the Combiner, and $p(\vh_{t}|\vx_{t})$ is estimated by the latent causal model, $p(\vv_{t+1},\va_{t+1}|\vh_t)$ by the modality reconstruction model (described below).
This allows for learning from previous representations in the sequence (in time) and aims to predict the next-step feature representation 
(\cref{fig:ar_arch}). 
While autoregressive modeling has been used for videos and images, it is often done on a pixel-by-pixel basis~\citep{weissenborn} which is highly inefficient and captures only short-term dependencies. With our approach, with autoregressive modeling and the Combiner, we address both shortcomings. 
We note that the Combiner also accumulates information from prior chunks, however, the autoregressive model works at a higher level of abstraction with already learned features from the Combiner. In the ablations, we find that it is most beneficial when both mechanisms work together. 

\textbf{Latent Causal Modeling.} The autoregressive latent model estimates:
$\prod_{t=1}^T p(\vh_t|\vx_t).$
This is done by applying an autoregressive transformer to $\bm{x} = \{\bm{x}_1, \bm{x}_2, \ldots, \bm{x}_T\}$ to produce $\bm{\hat{h}}=\{\bm{\hat{h}}_1, \bm{\hat{h}}_2, \ldots, \bm{\hat{h}}_T\}$ where the target of $\bm{\hat{h}}_t$ is $\bm{x}_{t+1}$ so the difference between $\bm{x}_{2, \ldots, T}$ and $\bm{\hat{h}}_{1, \ldots, T-1}$ is used as a loss to control the latent representation of the Combiner output $\bm{\hat{x}}$. Since we are modeling data autoregressively in time, care must be taken with the attention mechanism during training, the transformer uses a modified attention mechanism as described in~\cref{sec:timeattn},
~\cref{eq:timeattn}.

\textbf{Modality Reconstruction.} The autoregressive modality reconstruction models estimate
$\prod_{t=1}^T p(\vv_{t+1},\va_{t+1}|\hat{\vh}_t)$.
This is done by applying a separate transformer to $\bm{\hat{h}}$ to produce reconstructions of the audio and video signals $\bm{\hat{v}}$ and $\bm{\hat{a}}$, which is added as an optional loss below. To save on computation, the video input $\vv$ is down sampled to $\vv^{small}$ for the reconstruction target, thus the actual reconstruction is $\bm{\hat{v}}^{small}$.

\subsubsection{Attention mechanisms for Autoregressive modeling}
\label{sec:timeattn}

Since the autoreggressive models are trained in time, masking is done to satisfy causality. We note that the attention mechanisms within and across chunks need to be modified when masking. This applies to both the Combiner and the Autoregressive learning (Sections~\cref{sec:Combiner} and ~\cref{sec:autoregressive}). 
When masking features for autoregressive modeling, the standard pattern of masking each feature individually would mask features from within the same time-chunk from each other. While this would still satisfy causality, it unnecessarily restricts the model, preventing features from within the same time-chunk from interacting based on position within the time-chunk. To allow features in the same chunk to interact, the autoregressive mask between all features $\vi$, which fall in a time-chunk $t$, and another feature $\vj$ is computed as follows 
($N$ is the number of features and $T$ the number of time-chunks):
\begin{equation}
mask^i_j=\begin{cases}
      0 & j <= \text{ceil}(t*T/N)*N/T\\
      1 & \text{otherwise.}
    \end{cases}
\label{eq:timeattn}
\end{equation}


\subsection{Combining Aligned and Non-aligned Autoregressive Modeling}
\label{sec:autoregressive_text}

Text, or other context information, might not necessarily be aligned in time with the video and audio modalities. It is still sequential. So here it is modeled by a separate autoregressive model, devoted to text representations and to combining the visual-audio information together.
Assuming tokenization for the input text $\vt=\{ \vt^f_1, \vt^f_2, \ldots \vt^f_P \}$ is provided, obtaining a tokenized text sequence $\vt=\{ \vw_1, \vw_2, \ldots \vw_L \}$ of length L, we model the text sequentially as conditioned on audio and video. 
In order to combine the outputs of the video/audio autoregressive model we use cross-attention strategy~\citep{flamingo}. Here, unlike prior work, all feature representations $\hat{\vh}=\{\hat{\vh}_1, \hat{\vh}_2, \ldots, \hat{\vh}_T\}$ from the latent causal model are used in the main text model. 
\begin{equation}
p(\vw|\hat{\vh}) = \prod_{l=1}^L p(\vw_l|\vw_{l-1},\hat{\vh}).
\label{eq:text_ca}
\end{equation}
The autoregressive text model estimates~\cref{eq:text_ca} by applying a transformer to the input text sequence $\vw=\{\vw_1, \vw_2, \ldots, \vw_L\}$ and using the latent model output $\bm{\hat{h}}$ as cross-attention to produce $\bm{\hat{w}}$. The loss is the standard cross-entropy loss between target $\vw$ and output text sequences $\bm{\hat{w}}$. This provides further feedback to the Combiner latent representation $\hat{\vh}$ through the cross-attention layer. Of note is that since all parts of the model are autoregressive, it is naturally applicable to streaming videos.

\subsection{Model Losses}
\label{sec:learning}

We use two main losses, each driving the corresponding autoregressive model:
%

    \textbf{Latent space reconstruction loss} for time-aligned inputs is the difference between $\bm{x}_{2, \ldots, T}$ and $\bm{\hat{h}}_{1, \ldots, T-1}$ in the autoregressive setting such that $\bm{\hat{h}}_t=~\bm{x}_{t+1}$. We apply a $L^2$ normalization and then take dot product between the feature vectors as the loss (i.e., cosine similarity).
 
    \textbf{Unaligned text cross entropy loss} 
    is the standard cross-entropy loss between $\vw$ and $\bm{\hat{w}}$ for the unaligned text output.
    
    Additionally we implement the loss to encourage modality reconstruction as described in~\cref{sec:autoregressive}. More specifically we add a \textbf{video reconstruction loss} which is commonly used for video (audio reconstruction loss can also be added). Similar to latent space reconstruction above, the video reconstruction loss approximates the difference between $\bm{\hat{v}}^{small}$ and $\vv^{small}$ also in an autoregressive setting such that $\bm{\hat{v}}^{small}_t=~\vv^{small}_{t+1}$. We use the same distance metric on the video reconstruction as we use on the latent space reconstruction problem. While this loss can be useful, especially for generation tasks, we find that for our model, it is mostly subsumed by the latent space reconstruction loss. 
  %

%
These losses are weighted to compute the final loss. 

\begin{table}
\vspace{-0.2cm}
\begin{center}
\begin{tabular}{l|c}
Method	&Accuracy (\%)   \\
\midrule
\gray{Just Ask~\citep{yang2021justask}}	&\gray{41.5} \\ 
ALPRO~\citep{alpro}	&42.1 \\ 
\gray{MERLOT~\citep{merlot}}	&\gray{43.1} \\ 
VIOLETv2~\citep{fu2023empiricalmvm} &44.5 \\ 
VindLU~\citep{VINDLU} & 44.6 \\ 
VideoOFA~\citep{videoofa} &45.4 \\ 
GIT2~\citep{wang2022git} &45.6 \\ 
Iterative Co-Tok~\citep{piergiovanni2022cotok}	&45.7 \\ 
\gray{VideoCoca~\citep{videococa}}	&\gray{46.3} \\ 
All-in-one~\citep{wang2022allinone}  &46.8 \\ 
UMT-L~\citep{li2023umt} &47.1 \\ 
InternVideo~\citep{internvideo} &47.1 \\ 
Flamingo~\citep{flamingo}	&47.4 \\ 
M-PLUG2~\citep{mPLUG-2}	&48.0  \\ 
\midrule
\textbf{\ours - TTM} 	&\textbf{50.01} \\ 
\textbf{\ours} 	&\textbf{50.42} \\ 
\bottomrule
\end{tabular}
\caption{\textbf{Video QA results on MSRVTT-QA.} 
Results in gray show VideoQA as classification. 
}
\label{tab:video_qa}
\vspace{-7mm}
\end{center}
\end{table}

\subsection{Implementation details}
\label{sec:details}
\textbf{Model:} Our model has 3B parameters; without audio it is 2.9B. A little over half of the parameters are for the audio+video autoregressive model. Our models work on 128 frames customarily (16 chunks, 8 frames), but can handle more for longer videos (e.g., 512=16 chunks x 32 frames). We use Combiner dimension $m=32$. We apply random masking to the Combiner output features at a ratio of $0.75\%$ as a form of dropout regularization as we found this stabilizes the causal model latent reconstruction. 
%
Due to the design of our model (partitioning and Combiner), adding more frames, or increasing the chunk size, number of chunks, etc. lead to only marginal increase in parameters.  Increasing the number of chunks, while not leading to parameter increases, increases memory, which underscores the importance of the Combiner and particularly the TTM.   
%
%
\textbf{Model training:}
The model is pretrained on the Video-Text Pairs (VTP) dataset which is collected from noisy video-text pairs from the web~\citep{flamingo}. We use only about $12\%$ of the data, 3M samples.
%
%
%
All losses are given equal weight during pretraining.
%
%
%
%
%
During finetuning 
the unaligned text loss is increased 10-fold to better align the training loss with the final evaluation, which we also confirm experimentally. 



\vspace{-0.2cm}
\section{Experiments}
\vspace{-0.1cm}
We report results on standard Video Question Answering (VideoQA) benchmarks, on long-video VideoQA benchmarks and on Audio+Video benchmarks.
We report results using the \textbf{open-ended text-generative evaluation}, following 
~\citep{albef,singularity}. Our model generates a free-form text response which is compared to the target response for an exact match. This is more challenging than a classification setting, as our model might generate a correct answer (e.g. a synonym to the desired answer) but which is not among the target classes. 
This evaluation is more general and widely applicable. 

\textbf{Video Question Answering.}
We first report Video Question Answering results on the MSRVTT-QA VideoQA dataset~\citep{xu2016msrvtt}, as the most popular Video QA benchmark. 
The results are shown in~\cref{tab:video_qa} alongside the best state-of-the-art (SOTA) performances. 
Our method outperforms all prior methods on this challenging dataset, including the ones with classification evaluation which are at an advantage during evaluation. At less than 3B parameters, our model also outperforms the 5B GIT2~\cite{wang2022git} by a large margin, and outperforms the very big Flamingo~\citep{flamingo} of 80B parameters (the full fine-tuning Flamingo result~\citep{flamingo} is reported for direct comparison). 
This shows the benefit of our model design, where these results can be achieved with much fewer parameters and respectively much less compute needs.


\textbf{Long Video Question Answering.}
We further report Video QA results on long video datasets. ActivityNet-QA~\citep{Activitynet-qa} contains longer videos of about 160 seconds per video. NExT-QA~\citep{NExT-QA} is targeting complex events with long videos of about 44 seconds. We sample up to 512 frames (e.g. 16 chunks of 32 frames each)  without increasing the model size. 
Results are in~\cref{tab:video_qa_activity},~\cref{tab:video_qa_next}, showing we outperform the SOTA  with both 128 and 512 frames, 
where clear improvements are gained from using more frames, and without any increase in model size. We also outperform with either using Transformer Combiner or TTM Combiner.

\begin{table}
\begin{center}
\begin{tabular}{l|c}
Method	& Acc \%\\
\midrule
\gray{Just Ask~\citep{yang2021justask}}	&\gray{38.9 }  \\
\gray{MERLOT~\citep{merlot}}	&\gray{41.4}  \\
\gray{FrozenBiLM~\citep{frozenbilm}} &\gray{43.2}   \\
\gray{VideoCoca~\citep{videococa}}	& \gray{56.1}   \\   
\midrule
Sing-Temp~\citep{singularity} &44.1   \\
VindLU~\citep{VINDLU} & 44.7 \\
UMT-L~\citep{li2023umt} 	&47.9   \\
\midrule
\textbf{\ours - 512 frames TTM} 	&\textbf{49.85}   \\ 
\textbf{\ours - 128 frames} 	&\textbf{48.25}   \\ 
\textbf{\ours - 512 frames} 	&\textbf{51.13}   \\ 
\bottomrule
\end{tabular}
\caption{\textbf{Long Video QA results on ActivityNet}.
 Gray is for classification setting.}
\label{tab:video_qa_activity}
\end{center}
\end{table}

\begin{table}
\begin{center}
\begin{tabular}{l|c}
Method	& (Acc \%)\\
\midrule
CLIP (single frame) & 43.7 \\
VQA-T \citep{yang2021justask} & 52.32  \\
AIO \citep{wang2022allinone} & 50.60 \\
ATP \citep{buch2022revisiting} & 54.3 \\ 
VGT \cite{xiao2022video} & 55.02  \\
MIST - CLIP \cite{gao2023mist} & 57.18 \\
HiTeA \cite{HiTeA} & 63.1 \\
\midrule
\textbf{\ours - 512 frames TTM} 	&\textbf{73.2}   \\ 
\textbf{\ours - 128 frames} 	&\textbf{68.2}   \\ 
\textbf{\ours - 512 frames} 	&\textbf{72.0}   \\ 
\bottomrule
\end{tabular}
\caption{\textbf{Long Video QA results on NExT-QA}.
}
\vspace{-8mm}
\label{tab:video_qa_next}
\end{center}
\end{table}

\begin{table*} 
\vspace{-0.3cm}
\centering
\hspace{-5mm}\subfloat[\footnotesize{\textbf{Kinetics-Sound}.\label{tab:ablation:kin_sound}}]{
\begin{tabular}{lc}
Method	& Acc. \% \\
\midrule
\gray{MBT~\citep{attention-bottlenecks} (A+V)}	&\gray{85.0}   \\
\midrule
\ours  (Sm, Video)	& 81.3 \\
\ours  (Sm, A+V)	& 85.0  \\
\ours  TTM (A+V) & \textbf{88.3} \\
\ours  (A+V) & \textbf{90.1} \\
\bottomrule
\end{tabular}}
\subfloat[\footnotesize{\textbf{VGG-Sound}.\label{tab:ablation:vggsound}}]{
\begin{tabular}{lc}
Method	& Acc. \% \\
\midrule
UAVM~\citep{UAVM}            &65.8     \\
MMT~\citep{mmt}            &66.2     \\
MAViL~\citep{Huang2022mavil}         &67.1     \\
ONE-PEACE~\citep{ONE-PEACE}             &68.2  \\ 
\midrule
\ours  TTM (A+V)           &66.4  \\ 
\ours  (A+V)           &\textbf{69.8}  \\ 
\bottomrule
\end{tabular}}
\subfloat[\footnotesize{\textbf{Epic-Sound.}.\label{tab:epic-sound}}]{
    \begin{tabular}{l|c}
       Method & Acc. \% \\
         \midrule
      SSAST\citep{epicsounds} & 53.47 \\
      ASF\citep{epicsounds}  & 53.75 \\
      \midrule
      \ours (Audio) & 62.4 \\
      \ours (Video) & 72.4 \\
      \ours TTM (A+V) & \textbf{79.4} \\
      \ours (A+V)& \textbf{78.2} \\
      \bottomrule
    \end{tabular}}
    \vspace{-0.1cm}
\caption{\textbf{Audio-Video results on Kinetics-Sound, VGG-Sound, and Epic-Sound.}}
\vspace{-5mm}
\label{tab:video_audio}
\end{table*}

\begin{table*}
\centering
\subfloat[\footnotesize{\textbf{Effects of proposed components}.\label{tab:ablation:modelparts}}]{
\begin{tabular}{lccc}
Model & Frames/Chunks   & Acc. \\
\midrule
Baseline & 32/4 & 41.5 \\
+ AR & 32/4 & 43.2 \\
+ Combiner & 32/4 & 42.1 \\
+ AR + Combiner & 32/4 & 44.7 \\  
+ Pretraining & 32/4 & 45.2 \\
+ AR + Comb. + PT & 32/4 & 47.9 \\
\bottomrule
\end{tabular}}
\hspace{8mm}\subfloat[\footnotesize{\textbf{Combiner types}.\label{tab:ablation:Combiner}}]{
\begin{tabular}{lcc}
{Combiner type}     & Fr./Ch.   & Acc. \\
\midrule
Perceiver            &32/4 &43.1  \\
Transf.+CLS          &32/4 &43.7 \\
Ours-Transf.         &32/4 &44.2  \\
Ours-TTM             &32/4 &44.8  \\  
\bottomrule
\end{tabular}}
\vspace{2mm}
\subfloat[\footnotesize{\textbf{Autoregressive model more frames}.\label{tab:ablation:arm}}]{
\begin{tabular}{lccc}
{Model}  & Frames/Chunks &Acc.\\
\midrule
Baseline    &64/1   & 41.8  \\
Ours-Autoreg.    &64/8   & 45.1  \\
Ours + BD  & 64/8 & 45.1 \\
Ours-Autoreg.    &128/8   & 45.8  \\
\bottomrule
\end{tabular}}
\hspace{0.5cm}\subfloat[\footnotesize{\textbf{Combiner dimension}.\label{tab:ablation:Combiner_dim}}]{
\begin{tabular}{lccc}
{Model}     & Fr./Ch. & Dim   & Acc. \\
\midrule
Ours-8         &32/4    &8 &42.53   \\
Ours-16         &32/4    &16 &43.36   \\
Ours-32         &32/4    & 32 &44.20  \\
Ours-64         &32/4    & 64 &44.22  \\
\bottomrule
\end{tabular}}




\vspace{-0.1cm}
\caption{\textbf{Ablation studies on the MSRVTT-QA dataset.}} 
\label{tab:ablations}
\vspace{-5mm}
\end{table*}

\textbf{Audio-Video Results.}
\cref{tab:video_audio} shows results on three Audio-Video benchmarks: Kinetics-Sound~\citep{LookListen}, VGG-Sound~\citep{VGGSound} and Epic-Sound~\citep{epicsounds}. Since these datasets are Audio-Video classification, 
we treat the task as \textbf{open-ended generation}: 
 we input the text `Classify the video audio clip.' and expect the output to be the target class name e.g., `playing drums', where only an exact match is counted as accurate answer. Across all datasets, we outperform the SOTA with large margins, despite the more challenging open-text generation evaluation, as opposed to classification used in all prior works. 

\subsection{Ablations}
\label{sec:ablations}
The ablations (\cref{tab:ablations}), are conducted with the video and text model in order to understand the main behaviors of this architecture. 
We use a smaller model and configuration and 2x fewer pretraining steps with the same batch size to save compute (see the supp. material for more details).


\textbf{Main model components:} We first study the effect of each component (\cref{tab:ablation:modelparts}). We find that on top of a baseline model, adding each part, the autoregressive (AR) model, the Combiner, and  pretraining, each individually help and the combination of all three further help.

\textbf{Combiner type ablations:} We compare Transformer-based (ours, CLS and Perceiver \citep{flamingo}) and TTM Combiners. The CLS-token Combiner appends $m$ learnable features to the end of the sequence and takes their values as the combined features after passing the whole sequence through the transformer. Our main Combiners are shown in~\cref{fig:ar_arch_comb}. We use the same settings for direct comparison.~\cref{tab:ablation:Combiner} shows that our proposed Combiners perform best. 

\textbf{Autoregressive modeling in time:}
We ablate the Autoregressive part of the model.~\cref{tab:ablation:arm} shows that
processing the video in chunks autoregressively in time is more advantageous than learning from the full video at once, with a large jump in performance (first two rows). Not only is our autoregressive model feasible for longer videos but it is also more beneficial for the same size inputs.
More frames per chunk contribute to the improvements (rows two and four). We also compare to a bidirectional (BD) model, finding that the performance is the same as the autoregressive portion. 



\textbf{Combiner size ablations.} We compare the number of features output by the Combiner per timestep. We noticed a trend for larger Combiner outputs giving better results, lines 3-4 
(\cref{tab:ablation:Combiner_dim}). We chose 32 as a trade-off between sufficiently compact feature length and sufficiently expressive.

\vspace{-0.2cm}
\section{Conclusions}
 We propose a multimodal autoregressive model which decouples the autoregressive modeling  into a component, devoted to time-aligned modalities (video, audio) and another one for the non-aligned, contextual 
 modalities (text).  
 To address long video/audio inputs 
 we partition the media inputs and learn from them jointly by a Combiner, which allows to control the sequence lengths. 
 The model can handle 512 frames, without increasing its size. 
 Our approach not only enables working with long videos effectively but  also outperforms SOTA, achieving gains over previous models. 

{
    \small
    \bibliographystyle{ieeenat_fullname}
    \bibliography{main}
}


\clearpage
\appendix

\section{Datasets details}
The following datasets have been used for evaluation in the paper:

MSRVTT-QA~\citep{xu2016msrvtt} is a popular Video QA dataset of about 10K  video clips and 243K question-answer pairs. It is derived from the MSRVTT dataet by automatic question-answer pairs and contains a certain level of noise. 
Videos are about 14 seconds in length, on average.

 
 ActivityNet-QA~\citep{Activitynet-qa} is a commonly used benchmark for understanding of longer videos. It contains 5,800 videos and 58,000 question-answer pairs. 
It has much longer videos which entail longer and more complex scenes. The video length is about 160 seconds per video on average.

NExT-QA~\citep{NExT-QA} dataset is also addressing long video understanding. It contains 5,440 videos and about 52K manually annotated question-answer pairs. The average length of the videos is 44 seconds.
Apart from questions related to descriptions and in the video, NExT-QA focuses on questions related to events and sequence of events within the video, e.g., causal (`Why' and `How' questions), and temporal - questions related to order of events, or related to concurrent activities and others.

VGG-Sound~\citep{VGGSound} is a large-scale audio-video dataset, featuring over 200,000 videos accompanied by audio sounds. The data is formulated as classification tasks with 300 audio classes.

Epic-Sound~\citep{epicsounds} is an audio-video dataset based on the 
Epic-Kitchens dataset. It has 78.4k examples and 44 target classes.

Kinetics-Sound~\citep{LookListen} is a dataset derived from the popular Kinetics-400 video recognition dataset. Kinetics-Sound includes audio inputs sampled together with the video and has 36 classes.

All the abovementioned audio-video datasets used in the paper, have been formulated as datasets for classification tasks. Here we use the class outputs (which are typically short phrases describing an activity, instrument or type of sound e.g 'Knocking on a door') and treat them as open-ended text generation tasks and thus they are now audio-video-text datasets. 



\section{Additional ablations}

\cref{tab:ablations2} shows additional ablations. 
This is conducted by a model trained on only 1/2 of the epochs to save compute. All experiments within each ablation table are ran for the same steps.

\textbf{Autoregressive ablations, equalizing total dimensions}. In~\cref{tab:ablation:ar} we evaluate the autoregressive model vs non-autoregressive one, by equalizing the total number of Combiner dimensions. More specifically, if the full video is ran on $T$ chunks, each of Combiner dimension $K$, then we compare to a non-autoregressive model of total $T*K$ dimensions, in order to be maximally fair for both models. 
We see that, when equalizing the total dimensions, an autoregressive model is also more advantageous. More frames are beneficial, as expected, also confirming findings in the paper. We further see that allocating more dimensions, all other things being equal, is slightly beneficial. 

\textbf{Loss ablations:} We compare using different loss weights when training
(\cref{tab:ablation:text_loss}). 
We see that increasing the weight for the text generative loss is overall beneficial. This is done only during fine-tuning.
This ablation informed out decision to finetune the larger model using a larger unaligned text loss weight of 10.0.

\begin{table*}[t]
\centering
\subfloat[\footnotesize{\textbf{Autoregressive model}.\label{tab:ablation:ar}}]{
\begin{tabular}{lccccc}
{Model}  & Frames  & Chunks &Dim &Total Dim &Acc.\\
\midrule
Baseline    &32 &1 &256 &256 & 40.4  \\
Baseline    &128 &1 &256 &256  & 44.8  \\
Autoreg.    &128 &16 &16  &256 &45.5  \\
\end{tabular}}

\subfloat[\footnotesize{\textbf{Loss weights}.\label{tab:ablation:text_loss}}]{
\begin{tabular}{lcccc}
{Model}     & Causal & Video & Text   & Acc. \\
\midrule
Main         &1.0 &1.0 &1.0 &45.0 \\
Text Low             &1.0 &1.0 &0.1  &44.6   \\
Text High            &1.0 &1.0 &10.0  &45.4   \\
\end{tabular}}

\caption{\textbf{Additional ablation studies.}}
\label{tab:ablations2}
\end{table*}


\section{Combiner Visualizations.} 
\begin{figure}
    \centering
    \includegraphics[width=\linewidth]{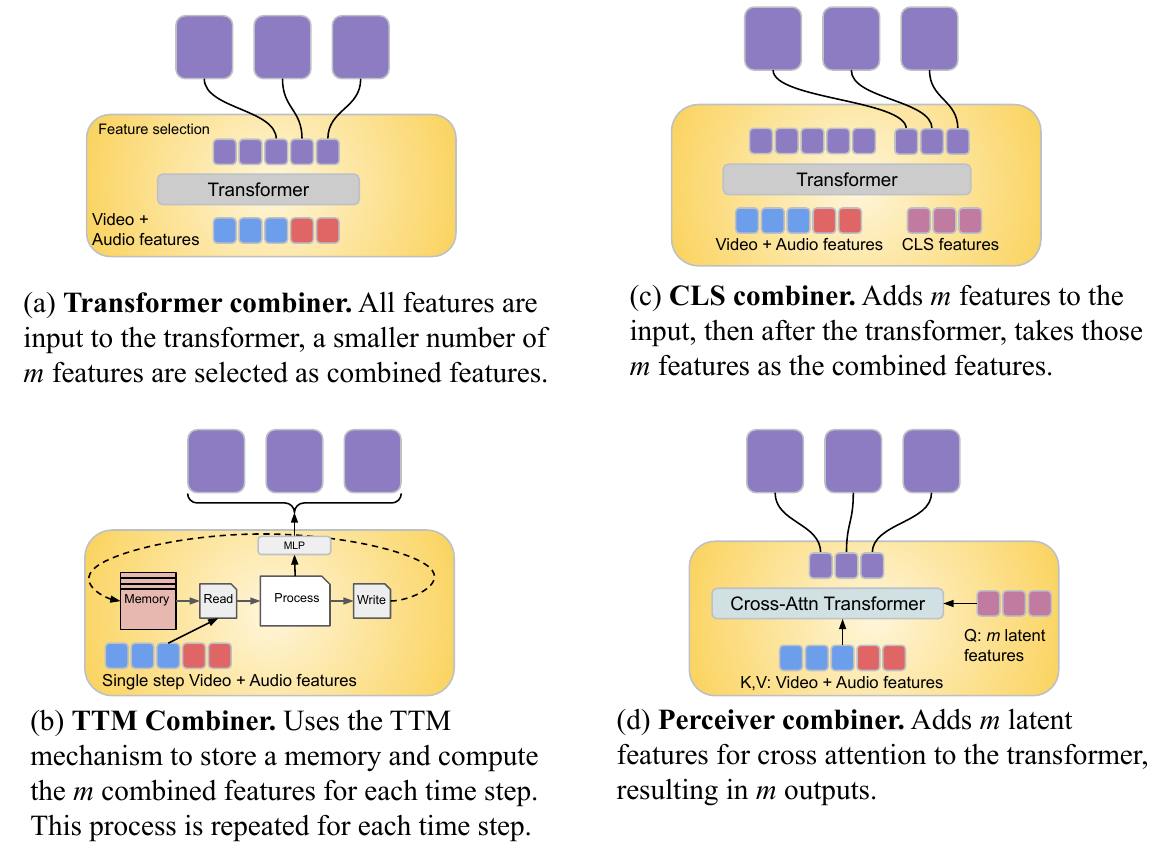}
    \caption{Visualization of the different combiners we explored in this paper. 
    The Transformer combiner, which is the main one we used, simply takes the last $m$ features of the output to represent the combined inputs. We found this to work well. The CLS combiner and Perceiver combiner we found both underperformed the base combiner. The TTM combiner is different, it uses a memory to store the previous representations and has read, process and write operations. We found this method saved memory with some tradeoff for accuracy for some datasets.}
    \label{fig:combiners}
\end{figure}

In Figure \cref{fig:combiners}, we visualize the different combiners we explored. The Transformer combiner, CLS combiner and Perceiver combiner are all based on transformers taking input of all the video + audio features and reducing them to $m$ combined features. We found our main combiner to outperform the other two in Table 5 of the main paper. 
We note that the Perceiver combiner is an adaptation of our combiner by applying Perceiver resampling~\citep{perceiver}.
The TTM combiner is conceptually different: rather than taking all the previous features as input, it takes only the current timestep features as input and uses a memory mechanism with read and write operations to update it. It then uses a MLP to produce the $m$ combined output features. This reduces memory and compute use and sometimes reduces accuracy.

\section{Additional Model and Implementation details}
\label{sec:impl_details2}

\textbf{Model Details.}
The autoregressive text model contains about 1.3B parameters, 400M are for cross-attention weights and 400M for the vocab embeddings and following specifications: layers=18, model dims=1536, hidden dims=12288, heads=12, and head dims=128. About 100M parameters are for the additional weights associated with audio. The remaining parameters are for the video input processor, combiner, causal latent model and video reconstruction model (a bit over 1.5B parameters in total). The combiner, causal latent model and video reconstruction model are transformers with 128M parameters and the following specifications: layers=8, model dims=1024, hidden dims=4096, heads=16, and head dims=64. The video chunk processor has roughly 630M parameters, following ViT-Huge. The convolutional tubes have 1.5M parameters and the transformer has 630M parameters and following specifications: layers=32, model dims=1280, hidden dims=5120, heads=16, and head dims=80. The total parameter size is 3B 
parameters.

The smaller model used for ablations keeps the same combiner, causal latent model, and video reconstruction model as the main model. However the autoregressive text model is reduced to 128M parameters with the same settings as the combiner, and has 20M cross-attention weights and 260M parameters for the vocab embedding. The audio parameters are held roughly the same. The video input processor is reduced to ViT-Large which has 300M parameters and the following specifications: layers=24, model dims=1024, hidden dims=4096, heads=16, and head dims=80. The total parameter size is 1.15B parameters.


The TTM Combiner, as mentioned is implemented by a TokenLearner~\citep{ryoo2021tokenlearner_neurips} function and a transformer. The output dimension $K=32$ is the same as the output dimension for the standard Transformer Combiner. The output dimensions for the `Read' and `Write' functions are 512 and 256, respectfully. These two parameters can be controlled independently to allow more or less capacity to the TTM Combiner.
The transformer used within the `Process' function is of 2 layers, 128 hidden dimension and 12 heads. These are fixed throughout the paper.

\textbf{Model Pretraining.}
The pretraining data is the Video-Text Pairs (VTP) dataset which is collected from noisy video-text pairs from the web~\citep{flamingo}. The main pretraining is done for the autoregressive, combiner, and the learning components processing the low-level video features (e.g., video tubes convolutions). The text backbone is frozen during pretraining while the other components including the cross attention weights are unfrozen. 
The model's image and text backbones and cross attention layers are initialized from a contrastively image-text pretrained MaMMUT model~\cite{MaMMUT}. More specifically, MaMMUT is trained jointly with contrastive and text generative objectives, where the latter is not of significant importance, and contrastive-only training is also possible. Pre-training is done on the Align dataset~\cite{align}. The audio backbone is also reusing the same pre-trained image backbone. 
%
%
During pretraining, the combiner model, causal latent reconstruction model and video reconstruction model and video tubes are all randomly initialized. All losses are given equal weight during pretraining. For pretraining, we used a learning rate of $1\times 10^{-5}$, batch size of 32, image resolution of $224\times 224$, 128 frames.
%
%
%
%

\textbf{Fine-tuning.} 
During finetuning all parameters are unfrozen. In addition the unaligned text loss is given extra weight and increased 10-fold to better align the training loss with the final evaluation, since the latent space and video reconstruction are not evaluated. The model is trained for 10 epochs for the MSRVTT-QA dataset and for 80 epochs on ActivityNet-QA and 20 epochs on NExT-QA. For these datasets, we finetune with a learning rate of $5\times 10^{-6}$, weight decay of $0.01$, 
image resolution of $448\times 448$, batch size of 32. We use 128 frames for the main experiments, except for the long video benchmarks where we also report performance with 512. Sampling more frames from the other benchmarks is not porductive as they contain relatively short videos. We used dropout of 0.1, label smoothing of 0.2


\textbf{Video-Audio Implementation Details.}
Since the model is pretrained on VTP data, where most videos lack audio, we add a further audio pretraining step here. We use AudioSet-2M \citep{audioset} and train the model to output the text of the class names. In this step, we freeze the weights of the model, other than the audio weights, allowing the model to learn to handle the spectrogram inputs. During fine-tuning on the eval datasets, we fully train the model. During finetuning, we also use Mixup \citep{mixup}, specaugment \citep{specaugment}, dropout and label smoothing, following the settings of previous works (e.g., \citep{audiomae}). We use a learning rate of $1\times 10^{-5}$, with the Adam optimizer (default settings), weight decay of $0.0001$, cosine learning rate decay. We use an image resolution of $448\times 448$, batch size of 32, and 128 frames.

\textbf{Ablation experiments details.}
The ablation experiments in Tab. 5a, 5b, 5c, 5d of the main paper are conducted with
our small model. The Baseline in Tab. 5a of the main paper uses partitioning, as the rest of the approaches tested in
the table, and concatenation of the features to be maximally comparable to others.


The baselines in~\cref{tab:ablation:ar} use a single time chunk which turn off aligned autoregressive modeling. The different chunks and dimensions explore the relationship between the number of frames, and output size of the combiner (dim). None of the single time chunk settings achieve the same performance as including a an autoregressive representation even at the same total dimensionality.




\end{document}